\documentclass[conference]{IEEEtran}
\IEEEoverridecommandlockouts

\usepackage{url}
\usepackage{amsmath}
\usepackage{amssymb}
\usepackage{amsthm}
\usepackage{algorithm}
\usepackage{algorithmic}
\usepackage{color}
\usepackage{mathtools}
\usepackage{graphicx}
\usepackage{hyperref}
\usepackage{subcaption}
\usepackage{float}
\usepackage{lipsum} 
\usepackage{bm}
\usepackage{multicol}
\usepackage{booktabs}
\usepackage{array}
\usepackage{wrapfig}
\usepackage{enumitem}
\usepackage{cite}
\usepackage{caption}
\usepackage{subcaption}
\usepackage{epstopdf}
\input{mysymbol.sty}
 
\makeatletter
\def\blfootnote{\xdef\@thefnmark{}\@footnotetext}
\makeatother

\newcommand{\ours}{Meta-Backward}

\newcommand{\eye}{\boldsymbol{I}}

\newcommand{\task}{\mathcal{T}}
\newcommand{\loss}{\mathcal{L}}
\newcommand{\data}{\mathcal{D}}
\newcommand{\inp}{\mathbf{x}}
\newcommand{\out}{\mathbf{y}}
\newcommand{\learner}{f}

\newcommand{\bR}{\mathbb{R}}

\newcommand{\param}{{\bm{\phi}}}               

\newcommand{\prior}{{\bm{\theta}}}               

\newcommand{\fn}{\mathcal{L}}                  

\newcommand{\pgrad}{\nabla}

\newcommand{\datatr}{\data^{\mathrm{tr}}}
\newcommand{\datatest}{\data^{\mathrm{test}}}
\graphicspath{{figures/}{../figures/}{./}{../../figures/}}
\def\BibTeX{{\rm B\kern-.05em{\sc i\kern-.025em b}\kern-.08em
    T\kern-.1667em\lower.7ex\hbox{E}\kern-.125emX}}

\hypersetup{
 colorlinks=True,
 linkcolor=blue,
 citecolor=blue,
 urlcolor=blue}

\makeatletter
\def\blfootnote{\xdef\@thefnmark{}\@footnotetext}
\makeatother

\begin{document}

\title{Energy-Efficient and Federated Meta-Learning via Projected Stochastic Gradient Ascent}
\author{\IEEEauthorblockN{Anis Elgabli, Chaouki Ben Issaid, Amrit S. Bedi$^\dagger$, Mehdi Bennis, Vaneet Aggarwal$^*$}
\IEEEauthorblockA{Centre for Wireless Communications (CWC) University of Oulu, Finland\\
$^\dagger${US Army Research Laboratory, MD, USA} \\
$^*$School of Electrical and Computer Engineering, Purdue University\\
Emails: \{anis.elgabli, chaouki.benissaid, mehdi.bennis\}@oulu.fi, {amrit0714@gmail.com}, vaneet@purdue.edu}
}

\maketitle

\begin{abstract}
In this paper, we propose an energy-efficient federated meta-learning framework. The objective is to enable learning a meta-model that can be fine-tuned to a new task with a few number of samples in a distributed setting and at low computation and communication energy consumption. We assume that each task is owned by a separate agent, so a limited number of tasks is used to train a meta-model. Assuming each task was trained offline on the agent's local data, we propose a lightweight algorithm that starts from the local models of all agents, and in a backward manner using projected stochastic gradient ascent (P-SGA) finds a meta-model. The proposed method avoids complex computations such as computing hessian, double looping, and matrix inversion, while achieving high performance at significantly less energy consumption compared to the state-of-the-art methods such as MAML and iMAML on conducted experiments for sinusoid regression and image
classification tasks.
\end{abstract}


\section{Introduction}
In many artificial intelligence problems, the aim is to develop systems that use previous experience to acquire new skills more effectively. To fulfill this objective, meta-learning has been proposed as a framework that enables an agent to quickly learn a new task by leveraging an optimized initializer that is trained using similar tasks~\cite{schmidhuber1987, thrun, naik}. One way to formulate the meta-learning problem is to cast it as a bi-level optimization problem~\cite{maclaurin2015gradient, finn2017model, rajeswaran2019meta}, involving two loops: (i) an inner loop optimization that represents adaptation to a given task, and (ii) an outer loop optimization with the objective to train a meta-model. In~\cite{finn2017model}, the authors proposed the model-agnostic meta-learning (MAML) formulation, which is shown to perform well in terms of convergence and fast adaptation to new task. However, despite its convergence and fast adaptation properties, solving MAML formulation requires backpropagation through the inner optimization process. Consequently, it requires higher-order derivative (Hessian's computation) which imposes high computational cost and vanishing gradients. These issues limit the scalability of MAML especially when the local models are multiple SGD steps from the initializer. First-order MAML (FOMAML) \cite{finn2017model} and Reptile \cite{nichol2018} are two first-order MAML-type algorithms which avoid second-order derivatives. Though both algorithms reduce the computational cost, they both ignore the dependence of task-models on the meta-models.  

In \cite{rajeswaran2019meta}, MAML with implicit gradients~(iMAML) was proposed with the goal of reducing the computational cost of MAML by leveraging implicit differentiation. The authors derived an analytical expression for the meta-gradient that depends only on the solution to the inner optimization and not the path taken by the inner optimization algorithm. However, this expression involves matrix inversion that has been approximated by solving a problem that involves using conjugate gradient (CG) method. Though iMAML reduces the computation cost compared to MAML, it still involves high computation cost due to: (i) the two loop procedure, (ii) the hessian computation which could be computationally expensive for large models, and (iii) the matrix inversion procedure using CG. Due to these limitations, running iMAML in a distributed environment that involves local computation and communication using energy-limited IoT devices is not feasible. Moreover, iMAML assumes a large number of labeled tasks since at every iteration, it samples a batch of these tasks, which may not be practical in cases where there are only few tasks owned by different agents. Therefore, in this paper, we propose a distributed and energy-efficient meta-learning approach that overcomes these limitations and ensures efficient learning.

Standard federated learning solves the problem of $N$ workers trying to learn a shared global model with the aid of a parameter server (PS). Each worker is exposed to limited data samples and is not willing to share its data due to privacy and/or communication constraints~\cite{elgabli2021harnessing,9357490,chenlag}. The PS starts with an initial guess of the global model, shares it with all contributing workers where each worker performs a local computation step before sharing its output. The local computation step could be (i) calculating the gradient of the local function with respect to the global model using a randomly selected mini-batch~\cite{chen2016revisiting} or (ii) performing a number of local stochastic gradient descent (SGD) steps using the global model as initializer ~\cite{mcmahan2017communication}. Then the PS aggregates the outputs of all workers and performs one global step which is one SGD step using the aggregated gradients of all workers for (i), or model averaging for (ii). Under some assumptions, this leads to convergence to a global model shared across workers.     

Similarly, in federated (distributed) meta-learning, every task is owned by an agent and each iteration involves communication with the PS. At each iteration, every agent updates its local model and uploads it to the PS where the meta-model is updated in a global step and pushed to all agents. Since this involves local computation and communication energy consumption, it is crucial to minimize the computation cost and save energy for communication. Motivated by this, we propose a novel formulation that can be solved very efficiently without involving double loop and Hessian computation. Particularly, at every iteration, we show that our inner problem can be solved in a simple and closed form expression that involves one SGD step and projection. Experiments using sinusoid regression and image classification tasks show that our proposed approach significantly outperforms iMAML in terms of energy consumption.

The rest of the paper is organized as follows. In section~\ref{mamlReview}, we briefly revisit MAML and iMAML and explain their key steps at high level. In section~\ref{ourFormulation}, we describe our problem formulation in details, then in section~\ref{ourAlgo} we introduce our proposed approach. In section~\ref{sec:experiments}, we evaluate the performance of our proposed method and compare it iMAML on sinusoid regression and image classification tasks. Finally, we conclude the paper in section~\ref{sec:conc}.

\textbf{Notations:} Throughout this paper, $\|\cdot\|$ denote the Euclidean norm of a vector. The notations $\nabla f$, $\nabla^2 f$ stand for the gradient and the Hessian of the function $f$, respectively.




\begin{figure*}[t]
\centering
\includegraphics[width=0.99\linewidth]{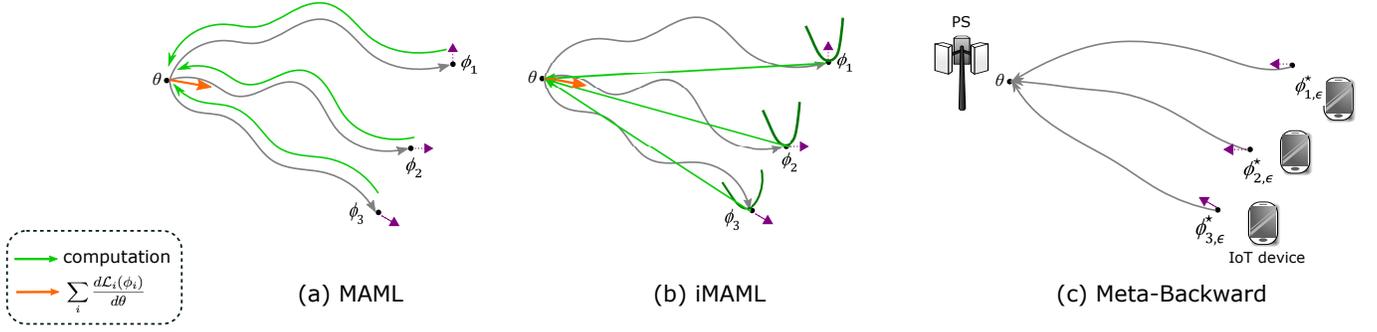}
\caption{Schematic illustration of our proposed algorithm ({\ours}) compared to MAML and iMAML.}
\label{fig:diagram}
\end{figure*}

\section{Review of MAML and IMAML} 
\label{mamlReview}
In this section, we will describe the concept and briefly highlight the key steps of MAML and iMAML.
\subsection{MAML}
Let $\{\task_i \}_{i=1}^N$ be a set of meta-training tasks assumed to be drawn from $P(\task)$. For each task~$\task_i$, we can sample two disjoint datasets: a training dataset $\datatr_i = \{ (\inp_i^k, \out_i^k) \}_{k=1}^{K}$ and a test dataset $\datatest_i = \{ (\inp_i^j, \out_i^j) \}_{j=1}^{J}$, where $\inp \in \mathcal{X}$ and $\out \in \mathcal{Y}$ denote the inputs and outputs, respectively. 

Introducing the loss function $\fn(\bm{\phi}, \data)$ where $\bm{\phi} \in \bR^d$, each task $\task_i$ aims to learn task-specific parameters $\bm{\phi}_i$ using $\datatr_i$ by minimizing its test loss, $\fn(\param_i, \datatest_i)$. MAML seeks to learn $\{\bm{\phi}_i\}_{i=1}^N$ using a set of meta-parameters $\bm{\theta} \in \bR^d$ and the training datasets from each task. The objective is to learn the meta-parameter that achieve good task specific parameters after adaptation. The meta-learning problem can be cast as the following bi-level optimization problem
\begin{align}
    \label{eq:objective}
    \min_{\prior, \{\bm{\phi}_i\}_{i=1}^N} \frac{1}{N} \sum_{i=1}^N \ \fn \left( \bm{\phi}_i(\prior), \ \datatest_i \right).
  \end{align}
Giving a new task $\task_i \sim P(\task)$, MAML can achieve a good generalization performance by using the adaptation procedure with the learned parameters $\param_i$ using $\prior$ and $\datatr_i$. More specifically, in the inner level of MAML, $\param_i$ can be updated using gradient descent initialized at $\prior$.
\begin{align}
    \label{eq:maml_gd}
  \param_i =  \prior - \alpha \nabla_\prior \loss(\prior, \datatr_i), 
\end{align}
where $\alpha$ is the learning rate. Note that MAML also allows updating $\param_i$ by running multiple gradient descent steps starting from $\prior$. However, a major concern with MAML is that updating the meta-model according to
\begin{align}\label{eq:outer_update_rule}
    \prior \leftarrow \prior - \eta \ \frac{1}{N} \sum_{i=1}^N \frac{d \bbphi_i (\prior)}{d \prior} \ \pgrad_{\bbphi} \fn(\bbphi_i(\prior), \datatest_i),
\end{align}
when $\bbphi_i(\prior)$ is $K$ SGD steps from $\prior$ requires backpropagating through the dynamics of the $K$ steps, which leads to a significant computation and memory burden.

\subsection{MAML with Implicit Gradients (iMAML)}\label{sec:proximal_regularization}
Due to requirements on memory and higher-order derivatives, MAML with Implicit Gradients (iMAML) was proposed in \cite{rajeswaran2019meta} to address the challenge of differentiating through the long optimization path in MAML. In addition to that, iMAML proposes to incorporate regularization into the problem formulation to enable appropriate leaning in the inner level, while preventing over-fitting. In this case, the inner-level problem is formulated as 
\begin{align}
    \label{eq:update_rule_supervised}
    \underset{\param}{\min}~\loss(\param, \datatr_i) + \frac{\lambda}{2}~||\param - \prior||^2,
  \end{align}
where $\lambda$ is a scalar hyperparameter that controls the regularization strength. The introduced regularization term in \eqref{eq:update_rule_supervised} enforces $\param_i$ to remain close to $\prior$. Ideally, the problem in \eqref{eq:update_rule_supervised} is solved using iterative algorithms and the optimal $\bbphi$ is determined. In this case, 
the term $d \bbphi_i (\prior)/d \prior$ can be explicitly derived as
\begin{align}
    \label{eq:alg_derivative}
    \frac{d \bbphi_i(\prior)}{d \prior} = \left( \eye + \frac{1}{\lambda}~ \pgrad_\param^2 \fn (\param_i, \datatr_i) \right)^{-1}.
\end{align}
According to iMAML, the problem in \eqref{eq:update_rule_supervised} is solved up to $\delta$-optimal solution, and the matrix inversion in \eqref{eq:alg_derivative} is determined by approximating a solution of an equivalent problem using a few conjugate gradient (CG) steps.

\section{Problem Formulation}
\label{ourFormulation}
In the standard MAML problem defined over $N$ different tasks, the goal is to find an initial point $\theta^\star$ such that when the local task $i$ runs its local $K$ SGD steps, it provides a better personalization for it. Hence, mathematically, we can write the problem of MAML as 
\begin{align}\label{mainBack03}
	\min_{\bbtheta, \{\bbphi_i\}_{i=1}^N}\frac{1}{N}\sum_{i=1}^N&\ \mathcal{L}(\bbphi_i^K(\bbtheta), \mathcal{D}_i^{test})\\
	\text{s.t.}\ \  
	\bbphi_i^K(\bbtheta) =& SGD_K^{\alpha}\left(\bbtheta,\mathcal{L}(\bbphi_i(\bbtheta), \mathcal{D}_i^{tr})\right), \forall i  \label{mainBack1_const102}
\end{align}
where $SGD$ denotes that stochastic gradient descent is used for each task with $\bm{\theta}$ as the starting point, i.e.,
\begin{align}
\bbphi_i^K(\bbtheta) =  \bbphi_i^0 - \alpha \sum_{k=0}^{K-1}\nabla_{\bbphi_i} \mathcal{L}(\bbphi_i, \mathcal{D}_i^{tr})\Big |_{\bbphi_i=\bbphi_i^k(\bbtheta)}, \forall i,
\end{align}
where $\bbphi_i^0 = \bbtheta$ and $\alpha$ is the learning rate. Now, let us see the importance of the starting point $\bbtheta$ for each task in more detail. The starting point actually controls the number of iterations required to converge to the local optima $\bbphi_i^\star(\bbtheta)$. If the local function is strongly convex, then the number of iterations required for $\epsilon$-optimal solution is given by \cite{bottou2018optimization}
\begin{align}\label{optimal}
	K_i\geq \mathcal{O}\left(\frac{\kappa}{\epsilon}\log\left(\frac{\mathcal{L}(\bbtheta, \mathcal{D}_i^{tr})-\mathcal{L}(\bbphi_i^{K_i}(\bbtheta), \mathcal{D}_i^{tr})}{\epsilon}\right)\right).
\end{align} 
Hence for each task $i$, the number of steps $(K_i)$ required to achieve $\epsilon$-optimal depends upon the term $\mathcal{L}(\bbtheta, \mathcal{D}_i^{tr})-\mathcal{L}(\bbphi_i^{K_i}(\bbtheta), \mathcal{D}_i^{tr})$ controlled by the initialization $\bbtheta$ for each task $i$. The goal of standard meta learning is to learn a common initializer $\bbtheta$ such that for each task $i$, after running a number of SGD steps using few data samples, we are able to obtain a good solution. So the placement of the initializer $\bbtheta$ in the space of local solutions $\bbphi_i^\star(\bbtheta)$ is the task.

We take a one step back and try to look into the problem in a backward manner. For instance, if we are given the local $\epsilon$-optimal solutions $\bbphi_{i,\epsilon}^\star$ (note that they are independent of $\bbtheta$) for each task $i$, then {\bf what should be the initializer point $\bbtheta$ so that each $\bbphi_{i,\epsilon}^\star$ is just $K_i$ SGD steps away from $\bbtheta$?} Since the final number of steps to converge to local optimal depends upon $\mathcal{L}(\bbtheta, \mathcal{D}_i^{tr})-\mathcal{L}(\bbphi_{i,\epsilon}^\star, \mathcal{D}_i^{tr})$, the best initializer point can be obtained by solving the following optimization problem
\begin{align}\label{9}
	\min_{\bbtheta, \{\bbphi_i\}_{i=1}^N}\ \ & \frac{1}{N}\sum_{i=1}^{N}\left(\mathcal{L}(\bbtheta, D_i^{tr})-\mathcal{L}(\bbphi_{i,\epsilon}^\star, \mathcal{D}_i^{tr})\right)^2
	\\
		\text{s.t.}\ \  
	& SGD_{K_i}^{\alpha}\left(\bbtheta, \mathcal{L}(\bbphi_i(\bbtheta), \mathcal{D}_i^{tr})\right)=\bbphi_{i,\epsilon}^\star , \forall i  \label{mainBack1_const103}
\end{align}
where $\bbphi_{i,\epsilon}^\star$ is known. Solving \eqref{9}-\eqref{mainBack1_const103} is as hard as solving the original MAML problem. Motivated by the fact that $\bbphi_{i}^\star$ is known and computed offline for every agent, we propose an approximate formulation for \eqref{9}-\eqref{mainBack1_const103} that can be solved using the proposed greedy backward algorithm described in Section \ref{ourAlgo}, which is shown to significantly outperform iMAML as illustrated in Section \ref{sec:experiments}. Our formulated problem is given by
 \begin{align}\label{mainForm}
   \min_{\{\bbphi_i\}_{i=1}^N}\ \ &\sum_{i=1}^N\sum_{k=0}^{K-1}\left( \mathcal{L}(\bbphi_i^k)-\mathcal{L}(\bbphi_i^{k+1}+\alpha\nabla_{\bbphi_i^{k+1}} \mathcal{L}(\bbphi_i, \mathcal{D}_i))\right)^2\\
   \text{s.t.}\ \ & \left\| \bbphi_i^k-\frac{1}{N}\sum_{i=1}^N\bbphi_i^{k+1}\right\|^2 \leq \delta_k, \forall i, k=0,\cdots, K-1 \label{mainForm_const}\\
   & \bbphi_i^{K}=\bbphi_{i,\epsilon}^\star, \forall i  \label{eps}
\end{align}
Note that $\bbtheta=\frac{1}{N}\sum_{i=1}^N \bbphi_i^0$, and $\delta_0 \leq \delta_1 \leq \dots \leq \delta_{K-1}$. When $\delta_k\rightarrow 0$, $\bbphi_i^0 \rightarrow \bbtheta, \forall i$. 

Let's now intuitively explain the relation between the formulation in \eqref{9}-\eqref{mainBack1_const103} and the one introduced in \eqref{mainForm}-\eqref{mainForm_const}. Originally, starting from $\bbphi_i^{k+1}$, $\bbphi_i^{k}$ is computed as 
\begin{align}
\bbphi_i^{k} = \bbphi_i^{k+1} + \alpha \nabla_{\bbphi_i^{k}} \mathcal{L}(\bbphi_i, \mathcal{D}_i).
\end{align}
However, when moving in backward manner, the term $\nabla_{\bbphi_i^{k}} \mathcal{L}(\bbphi_i, \mathcal{D}_i)$ cannot be computed. To overcome this problem, we replace $\nabla_{\bbphi_i^{k}} \mathcal{L}(\bbphi_i, \mathcal{D}_i)$ with $\nabla_{\bbphi_i^{k+1}} \mathcal{L}(\bbphi_i, \mathcal{D}_i)$ which are close to each other under smoothness assumption. However, the objective in \eqref{9}-\eqref{mainBack1_const103} does not reflect the fact that all agents need to meet at a common initializer after $K$ backward steps. To solve this issue, we introduce constraint \eqref{mainForm_const} which enforces that the local models of all agents get closer to each other at each step. Finally, the constraint \eqref{eps} ensures that $\bbphi_i^{K}$ is the $\epsilon$-optimal solution of agent $i$ ($\epsilon$-optimal in the norm 2 sense). Note that $\bbphi_{i,\epsilon}^\star$ is computed offline at each agent $i$ using an itertative optimization solver such that
\begin{align}\label{phistar}
\|\bbphi_{i,\epsilon}^\star - \bbphi_{i}^\star\| \leq \epsilon.
\end{align}
\section{Proposed Algorithm: \ours}
\label{ourAlgo}

Fig. \ref{fig:diagram} illustrates the idea of federated meta-learning and highlights the key differences of our proposed approach compared to MAML and iMAML. We set  $\bbphi_i^{K}=\bbphi_{i,\epsilon}^\star$ and define $\Phi^k = \frac{1}{N}\sum_{i=1}^N \bbphi_i^{k}$. We run our proposed backward meta-learning algorithm to find the meta model.

At backward step $k$, we need to solve the following problem to find $\bbphi_i^{k}, \forall i$ 
 \begin{align}\label{third}
   \min_{\{\bbphi_i\}_{i=1}^N}&\sum_{i=1}^N\left( \mathcal{L}(\bbphi_i)-\mathcal{L}(\bbphi_i^{k+1}+\alpha\nabla_{\bbphi_i^{k+1}} \mathcal{L}(\bbphi_i^{k+1}, \mathcal{D}_i))\right)^2\\
   \text{s.t.}\ \
   & \| \bbphi_i-\Phi^{k+1}\|^2 \leq \delta_k, \forall i \label{third_const}
\end{align}
Intuitively, starting from $\bbphi_i^{k+1}, \forall i$,  we want to find $\bbphi_i^{k}$ in an increasing direction that is closer to the average across tasks (agents). Therefore, by running $K$ backward steps while decreasing $\delta_k$, we eventually converge to a common initializer (meta model) that approximates the solution of \eqref{mainForm}-\eqref{eps}. We start by defining the feasible set
\begin{align}
{\cal X}_i=\{\bbphi_i : \parallel \bbphi_i-\Phi^{k+1}\parallel^2 \leq \delta_k\}.
\end{align}
With that, we can rewrite \eqref{third}-\eqref{third_const} as follows:
 \begin{align}\label{third0}
   \min_{ \{\bbphi_i\}_{i=1}^N}\ \ &\sum_{i=1}^N\parallel \mathcal{L}(\bbphi_i)-\mathcal{L}(\bbphi_i^{k+1}+\alpha\nabla_{\bbphi_i^{k+1}} \mathcal{L}(\bbphi_i^{k+1}, \mathcal{D}_i))\parallel^2\\
   \text{s.t.}\ \   & \bbphi_i \in {\cal X}_i, \forall i \label{third0_const}
\end{align}
Defining the indicator function 
\begin{equation}\label{equ:fetchingPolicy}
I_{{\cal X}_i}(\bbphi_i)=\left\{\begin{array}{l}
0, \quad \text{ if  $\bbphi_i \in {\cal X}_i$}\\
+\infty,  \quad \text{ otherwise}\\
\end{array}\right. 
\end{equation}
we can rewrite \eqref{third0}-\eqref{third0_const} as
\begin{align}\label{main}
\min_{ \{\bbphi_i\}_{i=1}^N} \sum_{i=1}^Nf_i(\bbphi_i),
\end{align}
where the function $f_i(\cdot)$ is defined as
\begin{align}
\nonumber &f_i(\bbphi_i) \\
&= \parallel \mathcal{L}(\bbphi_i)\!-\!\mathcal{L}(\bbphi_i^{k+1}\!+\!\alpha\nabla_{\bbphi_i^{k+1}} \mathcal{L}(\bbphi_i^{k+1}, \mathcal{D}_i), \mathcal{D}_i\big)\parallel_2^2\!+\!I_{{\cal X}_i}(\bbphi_i).
\end{align}
Projection gradient descent can be used to solve~\eqref{main}. Interestingly, it can be show that by choosing
\begin{align}
\bbphi_i^{k,0}=\bbphi_i^{k+1}+\alpha\nabla_{\bbphi_i^{k+1}} \mathcal{L}(\bbphi_i^{k+1}, \mathcal{D}_i),
\end{align}
we do not need to iterate more, so by projecting $\bbphi_i^{k,0}$ into the feasible set ${\cal X}_i$, we find $\bbphi_i^k$ for each task $i$. The projection problem for each task $i$ is formulated as
 \begin{align}\label{proj}
   \min_{\bbphi_i^k}\ \ &\frac{1}{2}\parallel \bbphi_i^k-\bbphi_i^{k,0}\parallel^2\\
   \text{s.t.}\ \ &\parallel \bbphi_i^k-\Phi^{k+1}\parallel^2 \leq \delta_k \nonumber
\end{align}
The Lagrangian function of ~\eqref{proj} is given by
 \begin{align}\label{proLagj}
L(\bbphi_i, \Phi,\mu)=\frac{1}{2}\parallel \bbphi_i^k-\bbphi_i^{k,0}\parallel^2 + \mu (\parallel \bbphi_i^k-\Phi^{k+1}\parallel^2 - \delta_k).
\end{align}
Therefore, the K.K.T conditions of ~\eqref{proj} are
\begin{align}
(\bbphi_i^k-\bbphi_i^{k,0})+2\mu(\bbphi_i^k-\Phi^{k+1}) &= \mathbf{0} \label{kkt1} \\
\mu (\parallel \bbphi_i^k-\Phi^{k+1}\parallel^2 - \delta_k) &= 0 \label{kkt2} \\
\mu &\geq 0 \label{kkt3}
\end{align}
Solving \eqref{kkt1}-\eqref{kkt3} yields the following solution
\begin{itemize} 
\item If $\mu=0$, then $\bbphi_i^k=\bbphi_i^{k,0}$. 
\item If $\mu>0$, then 
\begin{align}\label{kkt4}
\parallel \bbphi_i^k-\Phi^{k+1}\parallel^2 = \delta_k.
\end{align}
Deriving $\bbphi_i^k$ from \eqref{kkt1} as a function of $\mu$, and substituting it in \eqref{kkt4} yilds $\mu$ that solves
\begin{align}\label{muEq}
\mu^2+\mu-\frac{1}{4\delta_k}\parallel\bbphi_i^{k,0}-\Phi^{k+1}\parallel^2+\frac{1}{4}=0.
\end{align}
\end{itemize}
To sum up, compared to the state of the art algorithms, the proposed algorithm is computationally efficient since it has a closed form update expression in each iteration avoiding double loop. Moreover, it does not require any computational expensive operations such as matrix inversion. Instead, it only requires computing the gradient once and performing projection. The detailed steps of the proposed algorithm are explained in Algorithm \ref{alhead}.
		\begin{algorithm}[t]
			{ 				
			\begin{algorithmic}[1]
					\STATE {\bf Input}: $N, K, f_i(\boldsymbol{\phi}_i) \ \text{for all} \ i$
					\FOR {$i=0,1,2,\cdots,N$}
					\STATE Find $\phi_{i,\epsilon}^\star$ such that \eqref{phistar} is satisfied
					\STATE Set $\phi_i^{K}=\phi_{i,\epsilon}^\star$
					\STATE Transmit $\phi_i^{K}$ to the PS
					\ENDFOR
					\STATE PS computes $\Phi^K=\frac{1}{N}\sum_{i=1}^N \phi_i^K$ 
					\FOR {$k = K-1,\cdots,1,0$}
					\STATE {\bf Agent: } Each task (agent) finds $\phi_i^{k}$ that minimizes \eqref{main}  given $\phi_i^{k+1}$ and $\Phi^{k+1}$. 
					\STATE {\bf PS: } computes $\Phi^k=\frac{1}{N}\sum_{i=1}^N \phi_i^k$
					\STATE \hspace{0.65cm} transmits $\Phi^k$ to all agents
					\ENDFOR
				\end{algorithmic}
				\caption{\ours: Energy-Efficient Federated Meta learning \label{alhead}}
			}						
		\end{algorithm}

\section{Evaluation}
\label{sec:experiments}
In this section, we evaluate the performance of our proposed algorithm (\ours) compared to iMAML \cite{rajeswaran2019meta}. We start by describing the evaluation environment and then we discuss the evaluation results.

\begin{figure*}[t]
    \centering
         \centering
         \includegraphics[width=\textwidth]{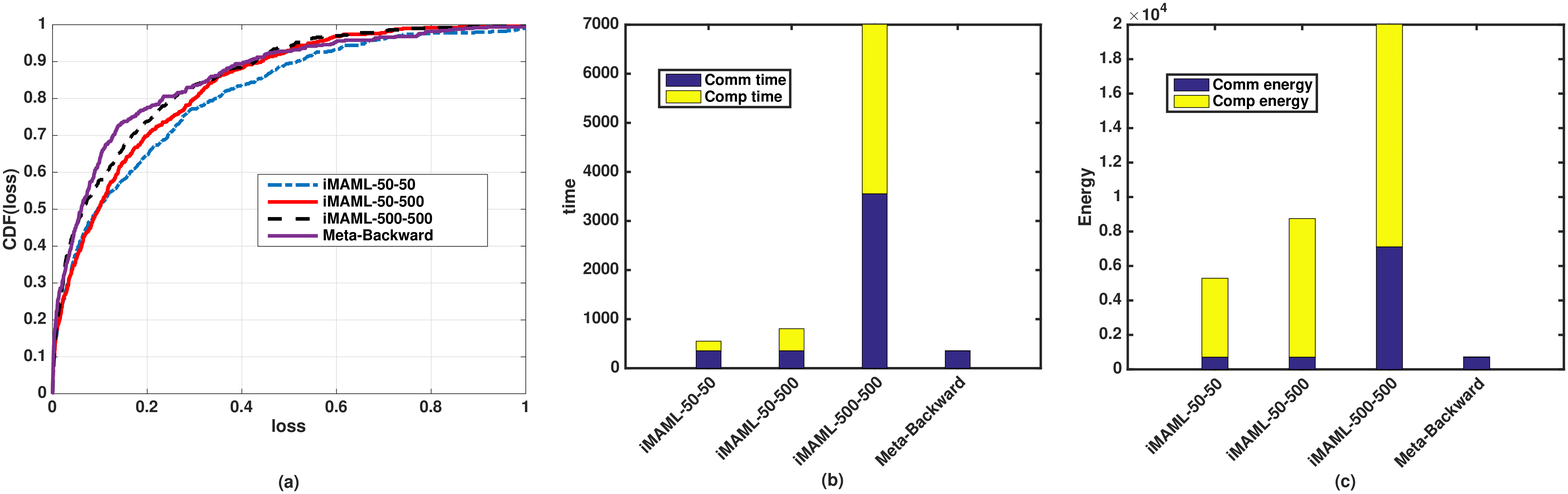} \\
             \caption{\footnotesize Accuracy, time, and energy of {\ours} and iMAML in the sinusoid example: (a) CDF of the loss, (b) Computation and communication time and (c) Computation and communication energy.}
    \label{fig:experiments}
\end{figure*}

\begin{figure*}[t]
    \centering
         \centering
         \includegraphics[width=\textwidth]{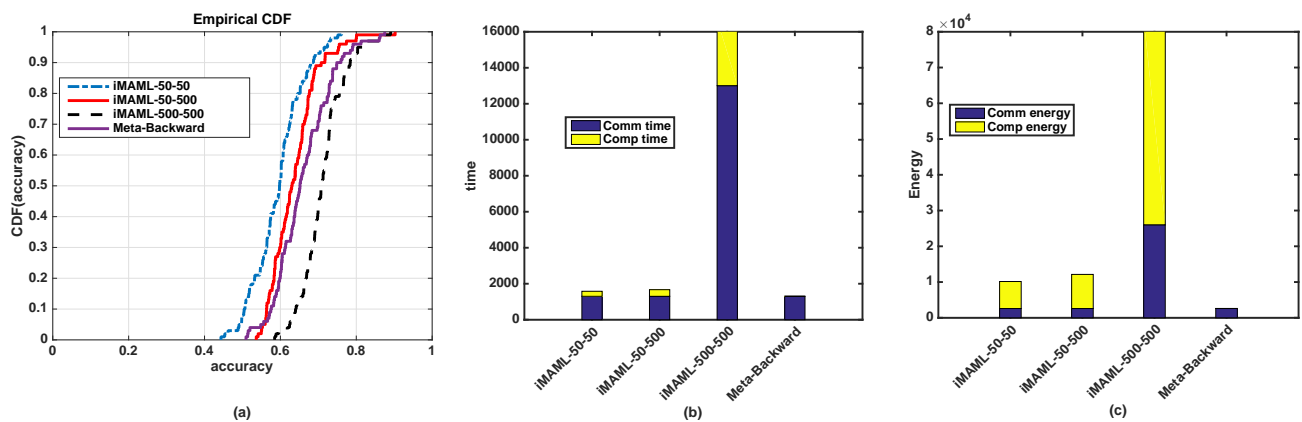} \\
             \caption{ Accuracy, time, and energy of {\ours} and iMAML in the image classification example: (a) CDF of the loss, (b) Computation and communication time and (c) Computation and communication energy.} 
    \label{fig:experiments2}
\end{figure*}

\subsection{Experimental settings}
{\bf Tasks: } We consider sinusoid regression and image classification tasks. For the regression problem, each task involves regressing from the input to the output of a sine wave. In particular, we train a meta model using 3 sinusoidal regression tasks with amplitudes $2$, $6$, and $10$, respectively, and then fine tune it to any sinusoid wave with amplitude ranges from $0.1$ to $10$ using $K$ samples. The loss is the mean-squared error between the prediction $\learner(\inp)$ and true value. The regressor we use is the same one used in \cite{finn2017model}, which is a neural network model with $2$ hidden layers of size $40$ with ReLU nonlinear activation functions. We choose $K=40$ for the sinusoid regression problem.

For the image classification task, we use the MNIST dataset \cite{LeCun:MNIST} comprising of 60000 training and 10000 test samples, each of which represents a hand-written $0-9$ digit image. We train using 2 tasks, where first task is trained to classify digits $0$ to $2$, and the second task is trained to classify digits $7$ to $9$. For both tasks, we use $400$ and $100$ samples for training and validation respectively. The two tasks are used to train a meta model that is then fine tuned to classify any $5$ randomly selected digits in the range $0$ to $9$ using $K$ samples from each un-seen class.  We choose $K=10$ for the classification problem considered in our experiment, so our problem is 5-way 10 shot classification. For the NN structure, we consider a 3-layer fully connected multi-layer perceptron (MLP) comprising an input layer with 784 neurons, two hidden layers with 8 neurons each, and an output layer with 10 neurons, resulting in the model size $d = 6442$. We use the rectified linear unit (ReLu) activation function, softmax output, and cross entropy loss.

{\bf Baselines: } We compare our proposed approach with iMAML under different settings where iMAML-$X$-$Y$ denotes running $X$ global SGD iterations to update the meta-model according to \eqref{eq:outer_update_rule}, and for each global iteration, each agent (task) runs $Y$ local SGD iterations to approximate the solution of the inner problem. For iMAML, we approximate the matrix inversion indicated in \eqref{eq:alg_derivative} using the same method described in \cite{rajeswaran2019meta} which involves $k$ conjugate gradient (CG) steps where we choose $k=5$ and $\lambda = 2$ as in \cite{rajeswaran2019meta}.

{\bf Performance metrics: } We consider the following metrics when comparing the performance of the proposed approach with the considered baselines.
\begin{itemize}
\item \textbf{Cumulative Distribution Function (CDF)} of the test loss/accuracy: After generating the meta-model of each algorithm, we fine-tune it to a new task using $K$ samples from each class. We randomly select a new task each time for $500$ and $100$ times for sinusoid regression and image classification problems respectively, and we report the CDF of the test loss/accuracy for all runs. As described earlier, we choose $K=40$ for the sinusoid regression problem and $K=10$ for the classification problem. Since there are $5$ classes in each new task for the classification problem, the total number of samples is $50$.

\item \textbf{Computation time and energy:} For the computation time, we use the time module in python, while the pyRAPL \cite{pyRAPL} package is used to evaluate the computation energy. pyRAPL measures the energy footprint of a host machine when executing a piece of Python code.
\item \textbf{Communication time and energy:} We assume communication over Additive White Gaussian Noise (AWGN) where the bandwidth is $B$ (hz) per agent, the transmission power is $P$ (W), and the noise power spectral density is $N_o$ (W/hz). Each element in the model is transmitted using $32$ bits. Hence, using Shannon formula, the transmission rate $R$ is
\begin{equation}
R=B \text{log}_2\left(1+\frac{P}{N_oB}\right).
\label{shannonCapacity}
\end{equation}

Then, we find the time ($\tau$) needed to send $32\cdot d$ bits, where $d$ is the model size. Knowing the transmission time and power, the communication energy per iteration and per agent is computed as $P \cdot \tau$. The total communication energy is then calculated by summing the per agent energy. The choices of $B$, $P$ and $N_0$ are specified for each example separately.
\end{itemize}

{\bf Framework and Hardware: } To run the experiments, we use Python 3.5 and TensorFlow 2.0 operated in a MacBook Air computer (1.8 GHz Intel Core i5 CPU, 8 GB 1,600 MHz DDR3 RAM).
\subsection{Results Discussion} 
Now, we describe our results for both sinusoid regression and image classification problems reported in Figs. \ref{fig:experiments} and \ref{fig:experiments2}.

\subsubsection{Sinusoidal Regression}

We first point out that we use 50 projected stochastic gradient ascent updates with a mini-batch size of $100$ samples and a fixed step size $\alpha=0.01$ for our proposed approach. For iMAML, we use the same choice for the mini-batch size and $\alpha$ for both the inner and outer loops. Note that we assume every task has been trained offline separately. To make the comparison fair, we use the average of the local models of all tasks as the initializer towards training the meta-model for all compared algorithms.

We evaluate the performance of all algorithms by fine-tuning the learned meta-model on $K=40$ data-points chosen randomly. In this experiment, we choose $B=$5Khz, $P=2$W, and $N_o=$1E-4, so the signal to noise ratio (SNR) is equal to $13.86$ dB. The qualitative results, shown in Fig.~\ref{fig:experiments}-(a)  show that {\ours} is able to achieve the minimum loss in more than $80\%$ of the experiments. In fact, iMAML requires more than 10 times local iterations (computation cost) and global iterations (communication cost) to approach the performance of \ours. Figs.~\ref{fig:experiments} (b) and (c) confirm these findings and show that {\ours} achieves this performance gain at significantly lower latency and energy compared to the other baselines. This can be explained by the simple and closed-form solution in each communication round performed by {\ours} compared to iMAML that requires at least 50 local iterations per communication round and involves complex operations such as matrix manipulation.

\subsubsection{Image classification}
In terms of communication setting, we use the same settings described for the sinusoid regression problem. As seen from Figs~\ref{fig:experiments2}-(a-c), the results confirm our findings from the sinusoid experiment highlighting that iMAML needs to pay significantly higher costs in terms of communication and computation energy to outperform our proposed approach. In other words, the proposed approach can achieve reasonable performance at very low energy cost using very few samples from the new task.

\section{Conclusion}
\label{sec:conc}
In this paper, an energy-efficient meta-learning approach has been proposed. The proposed approach enables learning a meta-model in a distributed setting and is shown to achieve high accuracy (low loss) for a randomly generated new similar regression/classification task with very few samples, at significantly less energy consumption compared to iMAML. The practicality of the approach was demonstrated using linear regression and image classification. Future work  will be focused on providing theoretical guarantees of the proposed approach in addition to investigating other use cases.

\bibliographystyle{IEEEtran}
\bibliography{main}

\end{document}